\documentclass[runningheads]{llncs}

\usepackage{eccv}

\usepackage{eccvabbrv}

\usepackage{graphicx}
\usepackage{booktabs}

\usepackage[accsupp]{axessibility}  

\usepackage[pagebackref,breaklinks,colorlinks,citecolor=eccvblue]{hyperref}

\usepackage{orcidlink}

\usepackage{listings}
\usepackage{multirow}
\usepackage{colortbl}
\usepackage{comment}
\usepackage{floatrow}
\newfloatcommand{capbtabbox}{table}[][\FBwidth]

\definecolor{mygray}{gray}{0.6}
\definecolor{mygray-bg}{gray}{0.9}
\definecolor{ablation_red}{RGB}{196,64,60}
\definecolor{ablation_green}{RGB}{0,155,85}

\def\OursData{Feint6K\xspace}
\def\OursMethod{LLM-teacher\xspace}

\begin{document}

\title{Rethinking Video-Text Understanding:\\Retrieval from Counterfactually Augmented Data} 

\titlerunning{Rethinking Video-Text Understanding: RCAD and Feint6K}

\author{Wufei Ma\inst{1,2}\thanks{All data collection and experiments were conducted at JHU.} \and
Kai Li\inst{1} \and
Zhongshi Jiang\inst{1} \and
Moustafa Meshry\inst{1} \and
Qihao Liu\inst{2}\\
Huiyu Wang\inst{3} \and
Christian H{\"a}ne\inst{1} \and
Alan Yuille\inst{2}}

\authorrunning{W.~Ma et al.}

\institute{Meta Reality Labs \and
Johns Hopkins University \and Meta AI}

\maketitle

\begin{abstract}
  Recent video-text foundation models have demonstrated strong performance on a wide variety of downstream video understanding tasks. Can these video-text models genuinely understand the contents of natural videos? Standard video-text evaluations could be misleading as many questions can be inferred merely from the objects and contexts in a single frame or biases inherent in the datasets. In this paper, we aim to better assess the capabilities of current video-text models and understand their limitations. We propose a novel evaluation task for video-text understanding, namely \textit{retrieval from counterfactually augmented data} (RCAD), and a new \textit{\OursData} dataset. To succeed on our new evaluation task, models must derive a comprehensive understanding of the video from cross-frame reasoning. Analyses show that previous video-text foundation models can be easily fooled by counterfactually augmented data and are far behind human-level performance. In order to narrow the gap between video-text models and human performance on RCAD, we identify a key limitation of current contrastive approaches on video-text data and introduce \textit{\OursMethod}, a more effective approach to learn action semantics by leveraging knowledge obtained from a pretrained large language model. Experiments and analyses show that our approach successfully learn more discriminative action embeddings and improves results on \OursData when applied to multiple video-text models. Our \OursData dataset and project page is available \href{https://feint6k.github.io}{here}.
\end{abstract}

\section{Introduction} \label{sec:intro}

Video-text foundation models have gained increasing attention due to their simple formulation and strong transferability \cite{ma2022simvtp,wang2022internvideo,li2023unmasked}. By pretraining on web scale video-text datasets, these models demonstrate strong performance across a wide range of downstream tasks, such as video-text retrieval \cite{xu2016msr,wang2019vatex} and video question answering \cite{xu2017video}.

As video-text foundation models evolve and achieve increasingly better performance on various benchmarks, we raise the following questions: Can these video-text model truly grasp the semantics of natural videos? Are these models genuinely reaching a level of understanding comparable to humans? Despite the remarkable achievements made in previous works, our study suggests that current video-text models still fall far behind human-level understanding.



\begin{figure}[tb]
  \centering
  \begin{subfigure}{0.4334\textwidth}
    \includegraphics[width=\textwidth]{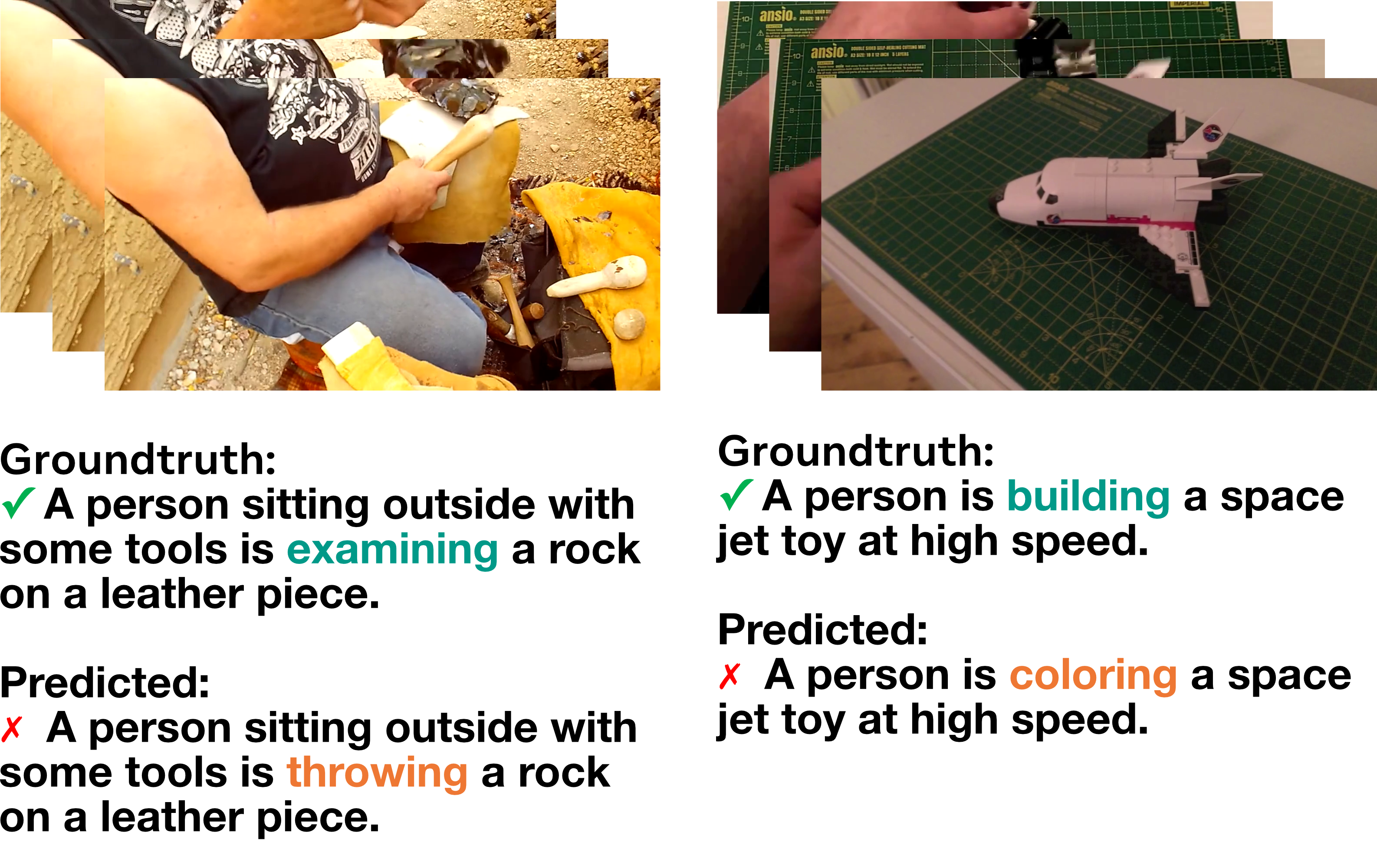}
    \caption{Failures of InternVideo \cite{wang2022internvideo} on \OursData}
    \label{fig:teaser_eg}
  \end{subfigure}
  \hfill
  \begin{subfigure}{0.5466\textwidth}
    \includegraphics[width=\textwidth]{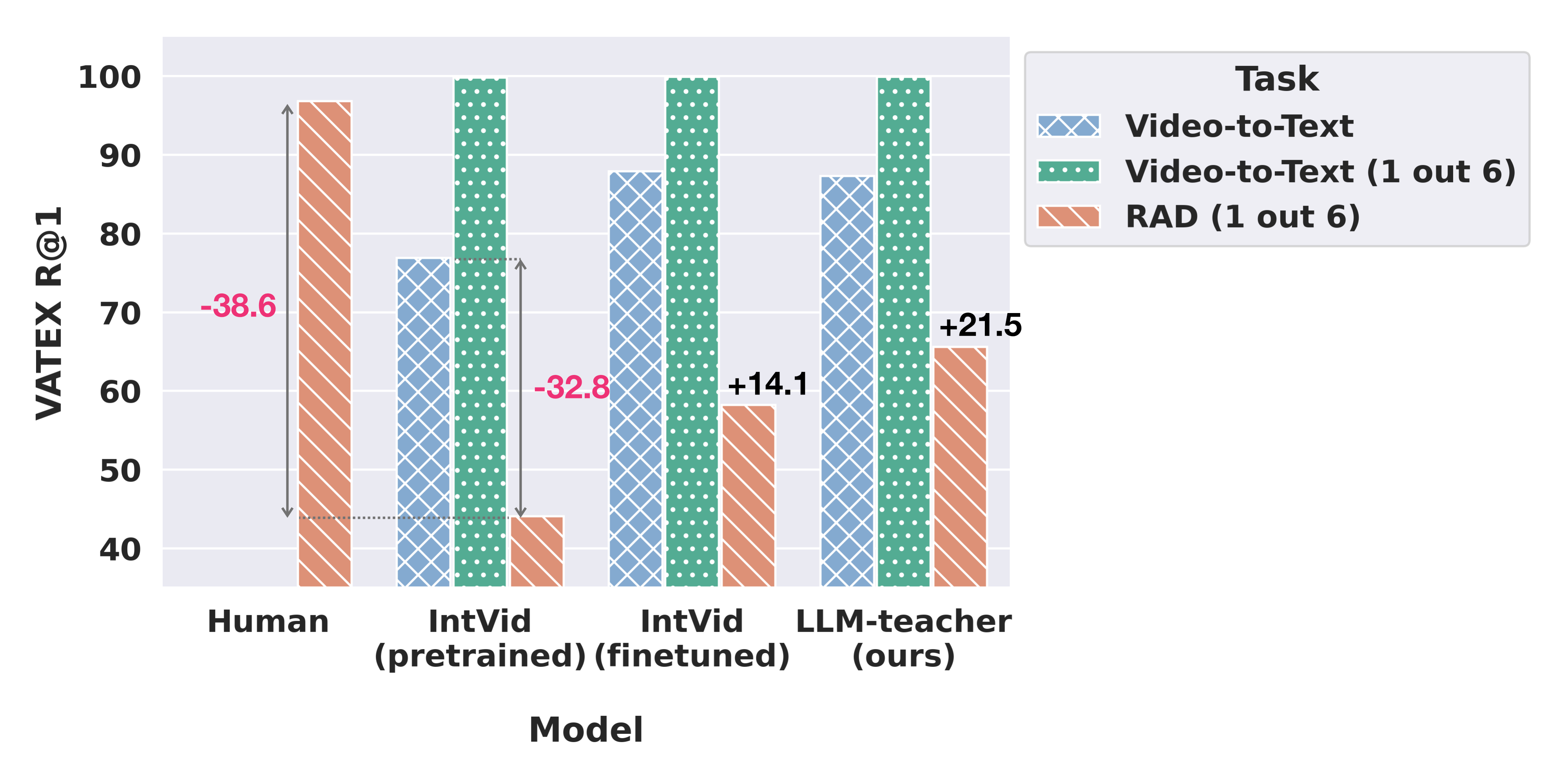}
    \caption{Quantitative results on \OursData dataset}
    \label{fig:teaser_compare}
  \end{subfigure}
  \caption{\textbf{(a):} Although with large-scale pretraining on web-scale data, current video-text model like \cite{wang2022internvideo} can be easily fooled by counterfactually augmented data. \textbf{(b):} The performance of InternVideo on \textit{retrieval from counterfactually augmented data} (RCAD) drops by over 30\% when compared to the standard video-to-text retrieval and by 38.6\% when compared to human-level performance. We also evaluate models on standard video-text retrieval from only 6 sampled candidates and show that our RCAD task is indeed much more challenging. Our \OursMethod successfully improves the performance on RCAD by enforcing a more effective learning of action semantics.}
  \label{fig:teaser}
\end{figure}

\begin{figure}[tb]
  \centering
  \begin{subfigure}{0.42\textwidth}
    \includegraphics[width=\textwidth]{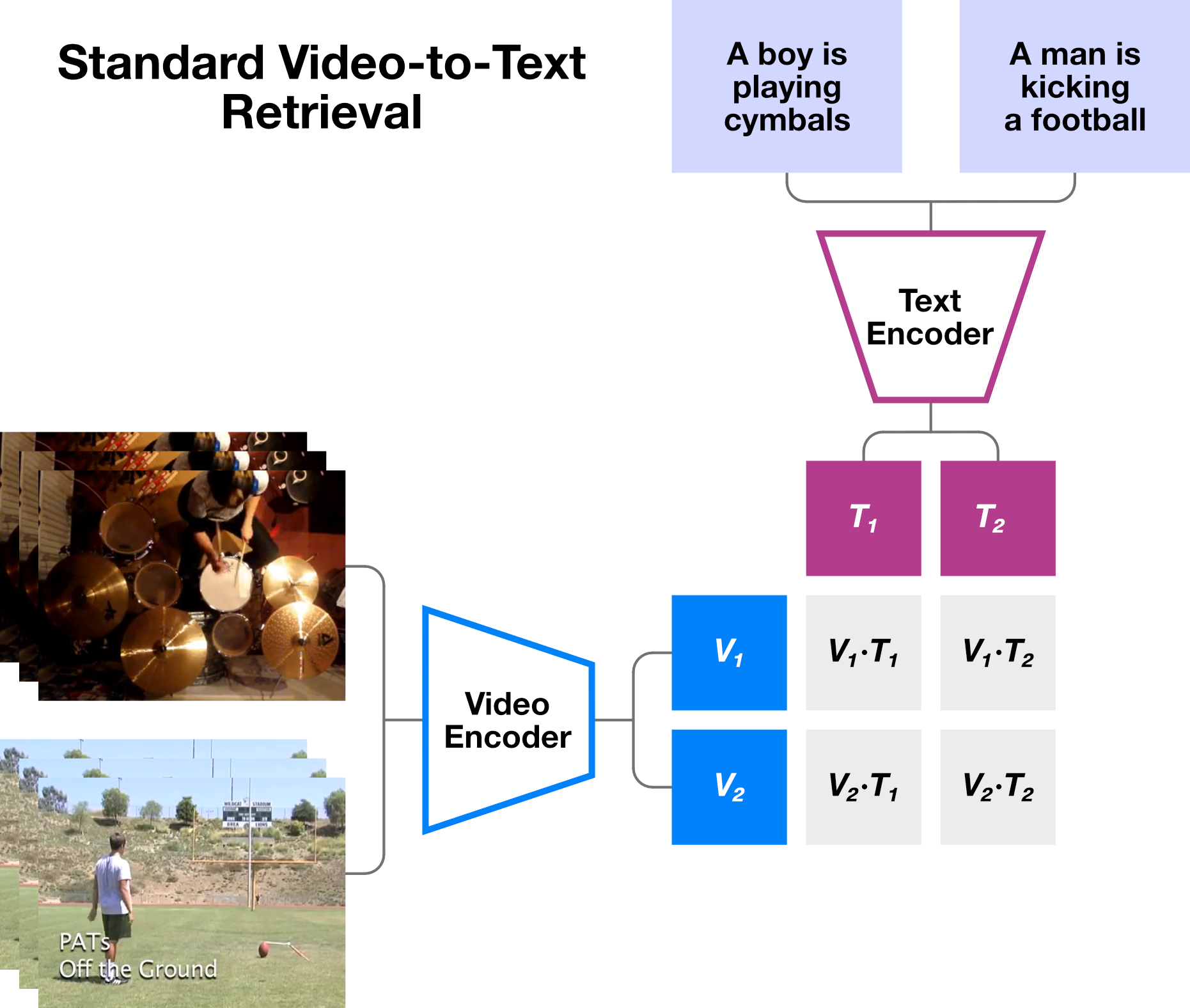}
    \caption{Standard video-to-text retrieval}
    \label{fig:standard_ret}
  \end{subfigure}
  \hfill
  \begin{subfigure}{0.42\textwidth}
    \includegraphics[width=\textwidth]{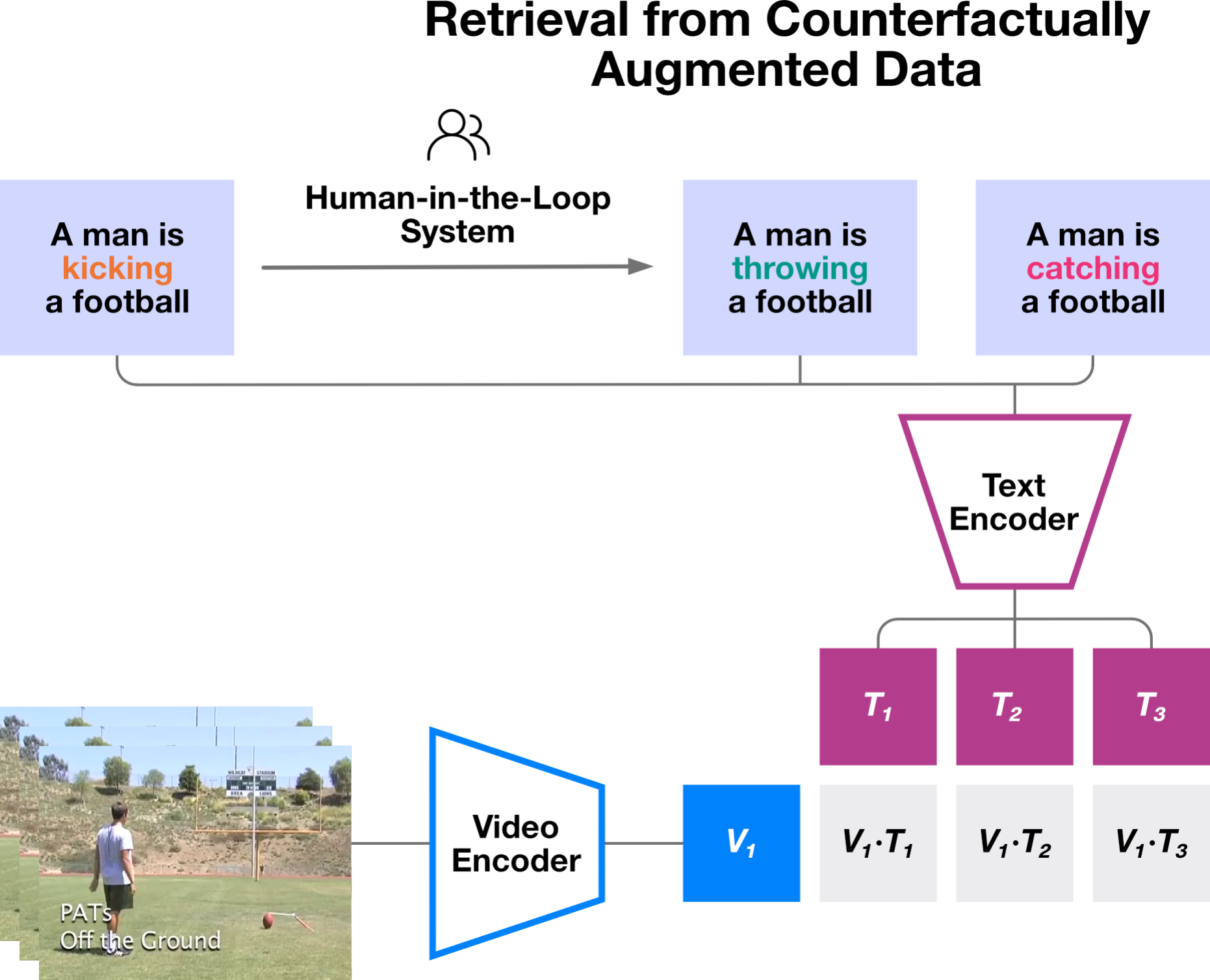}
    \caption{RCAD}
    \label{fig:ours_ret}
  \end{subfigure}
  \caption{\textbf{Different evaluations of video-text understanding.} \textbf{(a):} In standard video-to-text retrieval, negative captions are sampled from different videos in the same dataset. Image-text models can achieve good performance by exploiting shortcuts (\eg, ``football'' and ``cymbals'') and biases (\eg, spurious correlation between ``outdoor'' and ``football''). \textbf{(b):} In our proposed RCAD paradigm, we adopt a human-in-the-loop system (see \cref{sec:data_human}) to obtain ``hard'' negative captions with unchanged object entities but modified actions. Models must develop a holistic understanding of the semantics from the sequence of frames to retrieve the matched caption.}
  \label{fig:compare_ret}
\end{figure}

We argue that existing prominent results on standard video-text tasks can be misleading as models largely exploit the shortcuts and biases inherent in the dataset.
Many of the questions can be answered by objects or context extracted from a single frame without capturing cross-frame relations. For the examples in \cref{fig:standard_ret}, the video-text alignment can be easily inferred from shortcuts such as ``cymbals'' or ``football''. Moreover, the models may utilize biases in the datasets, such as the spurious correlation between ``outdoor'' and ``football''.
Current evaluations of video-text understanding are compromised by shortcuts and biases, which obscure us from analyzing the limitations of current models.
As we proceed from image-text understanding to video-text understanding, we should focus on more challenging semantics in the video domain that require cross-frame reasoning to solve, such as the interaction between person and object or the change of appearances over a sequence of frames (see \cref{fig:teaser_eg}).



In this paper we propose a novel evaluation paradigm of video-text understanding, \emph{i.e.}, \textit{retrieval from counterfactually augmented data} (RCAD). As demonstrated in \cref{fig:ours_ret}, the goal is to retrieve the only \textit{positive caption} with matched semantics among ``hard'' negative captions. \textit{Negative captions} are counterfactually augmented so that video-text models must derive a holistic understanding of the video sequence for both objects and actions, in order to retrieve the correct caption. As a comparison, negative examples from standard video-to-text retrieval \cite{xu2016msr,wu2017msvd,wang2019vatex} are captions of different videos in the same dataset, often containing different object entities that are easy to distinguish.

We follow the previous human-in-the-loop system \cite{kaushik2019learning} and develop a benchmark dataset for our new RCAD task, \emph{i.e.}, the \textit{\OursData} dataset. We test a wide range of public video-text methods with various pretraining strategies on this new benchmark. Two failure examples are visualized in \cref{fig:teaser_eg} and quantitative results are summarized in \cref{fig:teaser_compare}. Note how state-of-the-art video-text model, InternVideo \cite{wang2022internvideo}, features a 87.9\% rank-1 accuracy on standard video-to-text retrieval but drops by 32.8\% (pretraining only) and 29.7\% (finetuned) when tested on our benchmark with counterfactually augmented data. By establishing a human-level baseline on our benchmark, the InternVideo model surprisingly falls far behind human performance by 38.6\%. Our findings sharply contrast to the prominent performance obtained on previous video-text benchmarks and question the common belief that latest video-text models can develop a fairly effective representation for texts and videos in existing datasets.

In light of our evaluation results on \OursData, we identify a significant limitation of the widely adopted contrastive representation learning on video-text data, which is the issue of shortcut learning (see \cref{sec:method_shortcuts}). To address this, we propose \textit{\OursMethod} that enables a more effective learning of action semantics by introducing extra knowledge from pretrained large language models (LLMs). Specifically we generate synthetic captions by modifying the contents of an existing caption and LLM serves as the teacher to determine if a synthetic caption matches the semantics in the original caption using binary pseudo-labels or continuous logits. This approach enables a more effective learning of action embeddings from the available video-text data. Experimental results show that \OursMethod learns more discriminative action embeddings from our advanced contrastive objectives and improves the performance of retrieval from counterfactually augmented data when applied to multiple video-text models.

In summary, our key contributions are as follows.
(1) We propose to evaluate video-text models on questions that require cross-frame reasoning and develop a new task, retrieval from counterfactually augmented data, and a new dataset, \OursData (\cref{sec:data}).
(2) Extensive experimental results on our new evaluation paradigm suggest that existing video-text models demonstrate very limited understanding of the action semantics in a video, which is contrary to the prominent performance achieved on standard video-text retrieval benchmarks.
(3) From our results on \OursData dataset, we identify a key shortcoming of contrastive learning approaches on video-text data. We present \OursMethod to enforce a more effective learning of action embeddings by injecting knowledge from pretrained LLMs (\cref{sec:methods}). Our approach effectively improves the results on \OursData when applied to multiple text-video models.

\section{Related Works}

\subsubsection{Video-text pretraining.} With the availability of web-scale video-text paired datasets, such as WebVid2M \cite{Bain21} and HowTo100M \cite{miech19howto100m}, recent developments of video-text pretraining models achieved improved results on a wide range of video understanding tasks. Visual and textual embeddings jointly learned from the large-scale data demonstrate strong transferability and largely reduce the efforts in downstream tasks. Mainstream pretraining objectives can be categorized into discriminative and generative. Discriminative approaches extend the objective of CLIP \cite{radford2021learning} to the video-text domain and learn multi-modal representations by contrasting between matched and unmatched pairs \cite{xu2021videoclip,luo2022clip4clip,ma2022simvtp,wang2022internvideo} or simply predicting if a video-text pair is matched \cite{ma2022simvtp}. Generative methods follow the masked modeling idea in BERT \cite{devlin2018bert} and adopted masked language modeling (MLM) and masked video modeling (MVM) for video-text pretraining \cite{tong2022videomae,ma2022simvtp,wang2022internvideo}. These generative approaches are often considered superior for action recognition, while discriminative approaches learn better semantics from texts \cite{tong2022videomae}.

\subsubsection{Evaluation of video-text models.} In order to analyze the effectiveness of video-text pretraining, previous works \cite{wang2022internvideo,ma2022simvtp,xu2021videoclip,tong2022videomae,luo2022clip4clip,yan2022video} focused on the following tasks: (1) \textit{video-text retrieval (zero-shot or finetuned)} \cite{xu2016msr,wang2019vatex}: retrieving the matched video (or text) given a text (or video) as query where the negative candidates are unmatched pairs from the same dataset, (2) \textit{video question answering (finetuned)} \cite{xu2017video}: predicting the answer with a classification head, (3) \textit{video classification (zero-shot or finetuned)} \cite{kay2017kinetics,soomro2012ucf101,caba2015activitynet}: classifying videos using label names as text prompts, and (4) \textit{video captioning (zero-shot or finetuned)} \cite{xu2016msr,krishna2017dense}: summarizing the contents of a video.

We argue that results on these video-text tasks could be deceitful as most tasks heavily rely on the alignment of object entities in the video and text (which image-text models are also capable of), and ``action understanding'' can often be developed through shortcuts and biases. For instance, a commonly-used evaluation for ``action recognition'' is video classification on ActivityNet \cite{caba2015activitynet}, where a model classifies a given video into classes such as ``playing badminton'', ``kayaking'', or ``volleyball''. However, video-text models would classify the videos by exploiting objects and contexts as shortcuts (\eg, ``badminton'' and ``kayak''), as well as other biases (\eg, spurious correlation between ``kayak'' and ``on the water'')  without genuinely understanding the semantics of the action represented by interactions between the person and the objects over time. For instance, classifying the action as ``kayaking'' is essentially a ``kayak'' detection problem in ActivityNet. Therefore we propose \textit{retrieval from counterfactually augmented data}, a new evaluation paradigm where we aim to eliminate the shortcuts from the questions so models must establish a comprehensive understanding of the semantics from cross-frame reasoning in order to predict the correct answer.

\section{Retrieval from Counterfactually Augmented Data} \label{sec:data}

Previous evaluation tasks of video-text models focused on the alignment of feature embeddings of the video-text pairs (\textit{e.g.}, video-text retrieval \cite{xu2016msr,wang2019vatex} and video classification \cite{kay2017kinetics,soomro2012ucf101}) or reconstruction of the semantics with text (\textit{e.g.}, video question answering \cite{xu2017video} and video captioning \cite{xu2016msr,krishna2017dense}). However, these evaluation tasks are largely limited by the paired data available in existing video-text datasets. As demonstrated by in \cref{fig:standard_ret}, we can extract most of the semantics of a video, such as the objects, people, and contexts in the video, by looking at only one frame from the video. The action in the video can be further inferred by exploiting biases in the datasets. Therefore, models pretrained on image-text data often perform surprisingly well on existing video-text tasks.

To effectively evaluate how video-text models can understand the contents and semantics of videos beyond images, we must develop video understanding tasks that are free from shortcuts and require cross-frame reasoning to solve. To this end, we propose a new evaluation task for video-text models, \ie, retrieval from counterfactually augmented data (\cref{sec:task_intro}), and a new dataset \OursData (\cref{sec:data_human}). In contrast to previous datasets with matched video-text pairs scrapped from the web \cite{xu2016msr,wu2017msvd} or annotated by human \cite{wang2019vatex}, we adopt a human-in-the-loop system and annotate counterfactually manipulated texts as negative pairs with the original video. This allows us to evaluate video-text models with novel tasks and better understand their limitations. Lastly we measure human performance on our dataset as a direct comparison with the state-of-the-art video-text models (\cref{sec:human_performance}).

\subsection{Task Formulation} \label{sec:task_intro}

\begin{figure}[tb]
  \centering
  \includegraphics[width=\textwidth]{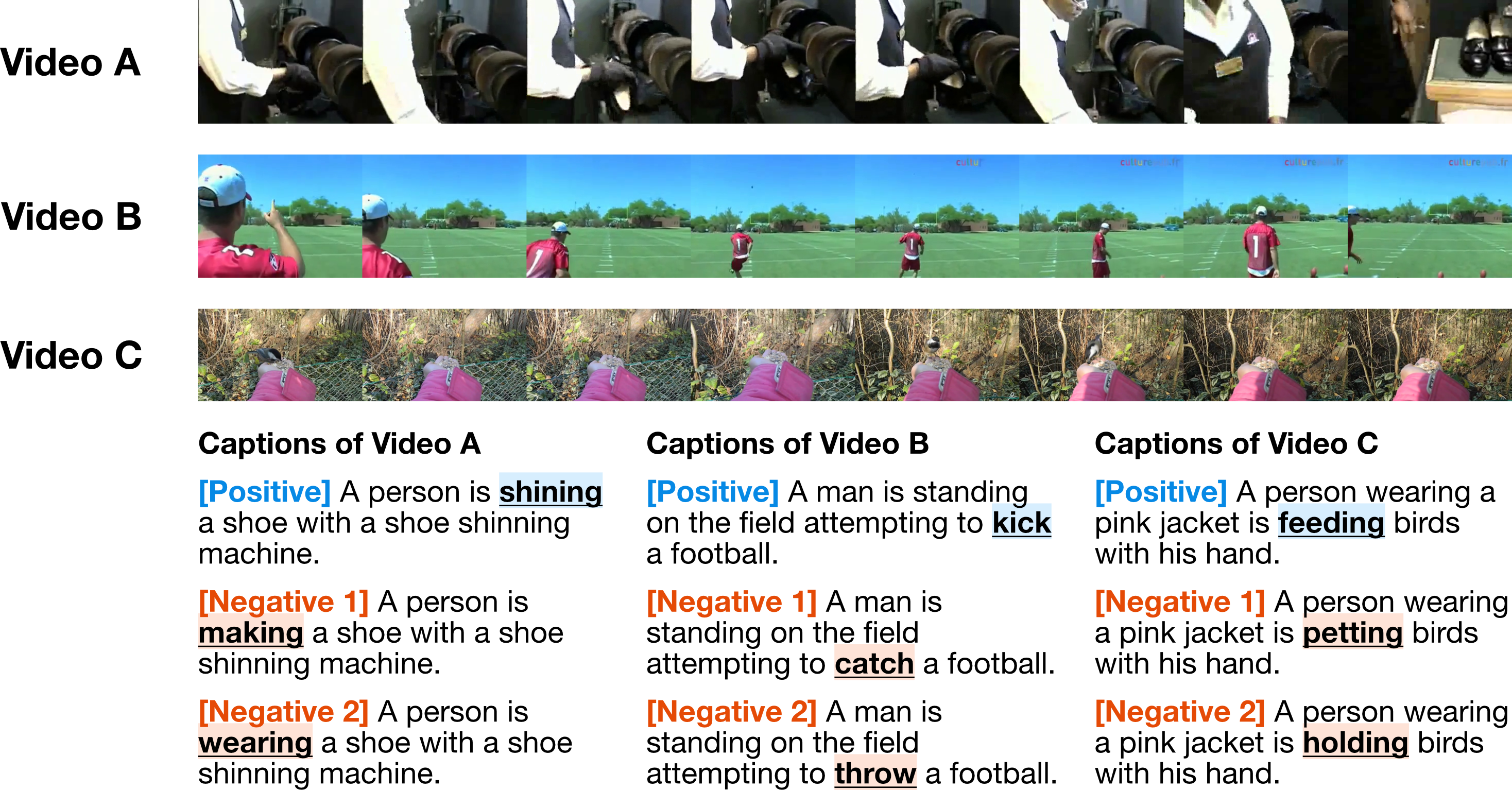}
  \caption{Examples of \textit{RCAD} on our \OursData dataset.}
  \label{fig:eg_task}
\end{figure}

Retrieval from counterfactually augmented data is a variant of the standard video-to-text retrieval. As shown in \cref{fig:ours_ret}, given a video sequence and a list of candidate captions, the model is asked to retrieve the caption that best matches the semantics in the video. However, unlike standard video-to-text retrieval where negative captions come from other video-text pairs in the same dataset, negative captions in our task are modified from the positive captions, with the same text structure and object entities but different actions (see \cref{fig:eg_task}).

Similar to video-text retrieval, our proposed task provides the benefit of zero-shot evaluation. This allows us to quantitatively analyze the effectiveness of video-text pretraining without downstream finetuning.

What distinguishes our task from all previous video-text tasks is its focus on questions that require cross-frame reasoning. Consider ``Video B'' in \cref{fig:eg_task} as an example. All captions contain the same object entities present in the video, \ie, ``a man'', ``the field'', and ``a football'', but differ in the action specified. We cannot determine the most accurate caption because all suggested actions, ``kick'', ``catch'', and ``throw'', are plausible given the video's context. To retrieve the corresponding caption, it is crucial to grasp the action's semantics by inspecting the interactions between the man and the football across a series of frames. This underscores the necessity of understanding the video semantics over time in order to succeed in our task, rather than relying shortcuts or biases.


\subsection{\OursData: Data Collection} \label{sec:data_human}

We utilize a human-in-the-loop system \cite{kaushik2019learning} to counterfactually manipulate the positive captions in existing video-text datasets \cite{xu2016msr,wang2019vatex}. The generated counterfactually augmented data, when paired with the original videos, forms negative pairs to be used in our proposed task. We recruit 40 annotators to manually manipulate the actions in existing captions. Specifically, the new actions introduced must be plausible within the context in the caption but are not occurring in the corresponding video. In summary, a total of 6,243 videos from the validation set of the MSR-VTT dataset \cite{xu2016msr} and the test set of the VATEX dataset \cite{wang2019vatex}. For detailed statistics of our \OursData dataset, please refer to our supplementary materials.

\begin{figure}[tb]
  \centering
  \includegraphics[width=0.9\textwidth]{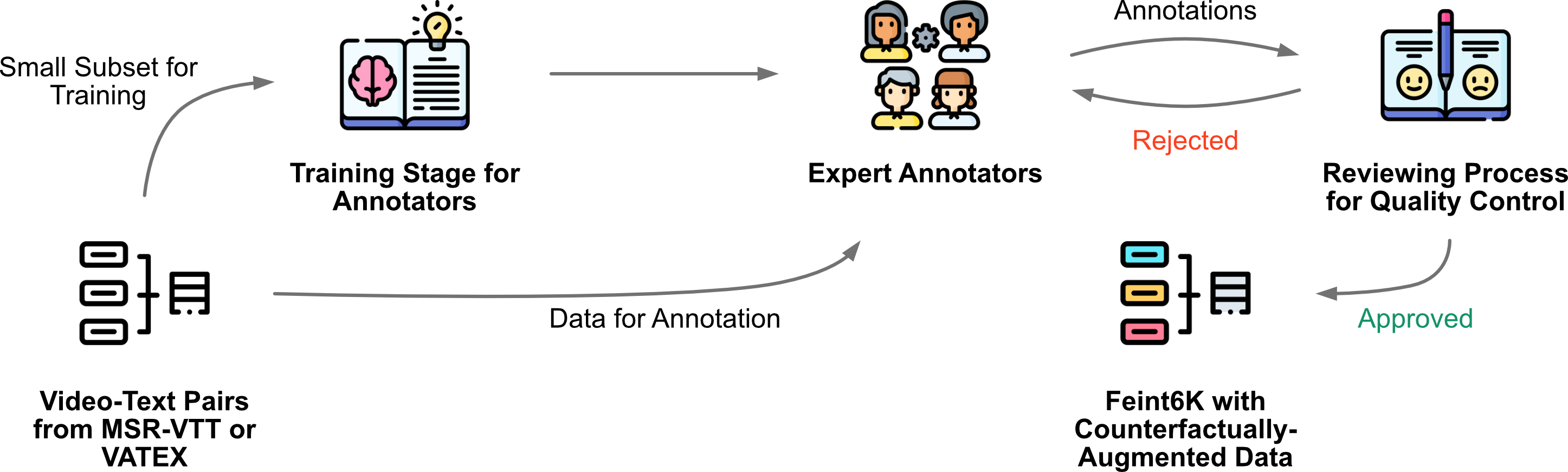}
  \caption{Overview of our data collection pipeline for \OursData dataset featuring a human-in-the-loop system.}
  \label{fig:hitl}
\end{figure}

The overview of our human-in-the-loop system is demonstrated in \cref{fig:hitl}. We start by sampling a small subset of video-text pairs and manually annotate the counterfactually augmented captions for the purpose of training and demonstration. Specific guidance and feedback are given to the annotators on the practice questions during the training stage. Moreover, we adopt a reviewing process during the annotation stage to further ensure the quality of our annotated data. Each annotation is reviewed before acceptance, where rejected annotations are marked and sent back for refinement. We refer the readers to the supplementary materials regarding the annotation interface and annotator guidance.

\subsection{Human Performance} \label{sec:human_performance}

We employ the same group of annotators to establish a human-level baseline on our \OursData dataset. Analyzing the human performance serves two goals: (i) verifying the legitimacy of our \OursData dataset -- if each question is answerable and has one unique answer, and (ii) comparing the state-of-the-art video-text models with human-level performance.

Each annotator is first presented with the full video and then asked to select the one out of six captions that best matches the semantics in the video. To avoid leakage, we make sure the annotators assigned to the testing questions are not the same annotators annotating this video. We also randomly shuffle the order of the candidate captions to remove any biases. In \cref{sec:main_results} we present the quantitative comparisons between human-level performance and current state-of-the-art video-text models on our \OursData dataset.

\section{\OursMethod} \label{sec:methods}

Contrastive learning are widely adopted in previous video-text representation learning works \cite{xu2021videoclip,ma2022simvtp,wang2022internvideo}. The contrastive models learn to distinguish between similar and dissimilar pairs of video-text data by contrasting their feature representations. Despite the simple formulation and great generalization ability, our results on \OursData show that current video-text models built heavily on contrastive objectives have very limited understanding of action semantics in natural videos and can be easily fooled by counterfactually augmented data. Therefore we aim to develop a more effective contrastive approach and learns more discriminative action representations from existing video-text data.


\cref{sec:method_shortcuts} discusses the limitations of current contrastive approaches due to numerous shortcuts in video-text dataset. Inspired by this finding, we present \OursMethod in \cref{sec:our_method}, a simple but effective approach to learn more discriminative action embeddings by incorporating knowledge from pretrained large language models.

\subsection{Shortcuts in Video-Text Data} \label{sec:method_shortcuts}

The CLIP model \cite{radford2021learning} trained on image-text data with a contrastive objective demonstrated strong zero-shot capabilities in many vision tasks \cite{zhou2023zegclip,lin2023gridclip} and was widely adopted as the vision encoder in many large vision-language models \cite{liu2023llava}. However, directly generalizing this idea to video-text data may face unprecedented challenges due to shortcuts in the video-text data.

As we proceed from image-text to video-text representation learning, we aim to learn a powerful embedding for not only the object entities, but also the actions specified by texts or sequence of frames. Naturally we hope to achieve this by contrasting videos with different actions. However, as we are optimizing the contrastive objectives in a mini-batch of videos, object entities become the shortcuts that contrastive models exploit to saturate the contrastive objective, leading to an ineffective action embeddings. Consider the example in \cref{fig:standard_ret}, the contrastive loss would be small as long as the model learns discriminative embeddings for shortcut objects such as ``cymbal'' and ``football'', even without any understanding of the actions. Since many video-text models start with pretrained CLIP encoders \cite{xu2021videoclip,wang2022internvideo}, they often posses strong embeddings for shortcut objects from the beginning of the video-text pretraining, which hinders the model from further learning effective action representations.

\begin{figure}[tb]
  \centering
  \begin{subfigure}{0.49\textwidth}
    \includegraphics[width=\textwidth]{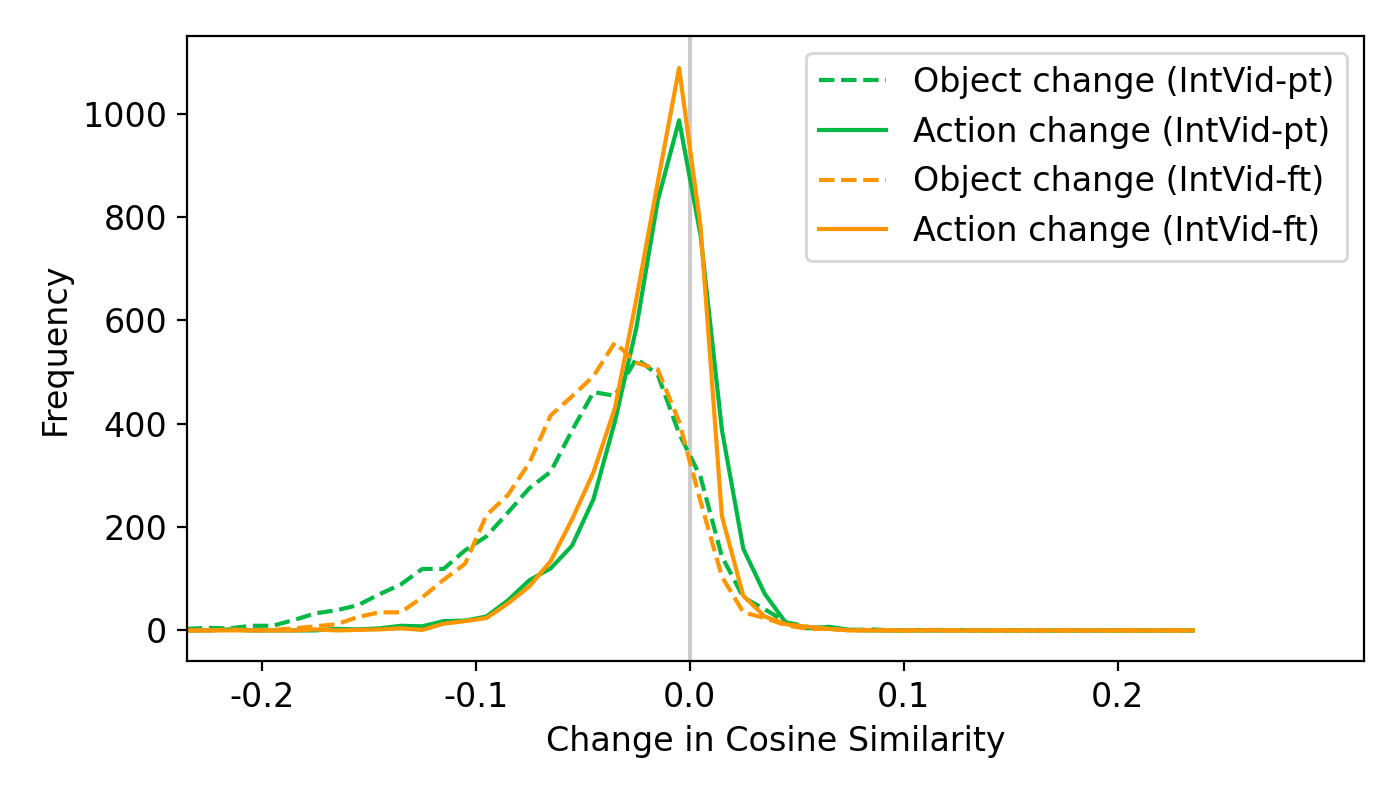}
    \caption{Comparison between $\Delta s$ \wrt object or action changes.}
    \label{fig:sense_1}
  \end{subfigure}
  \hfill
  \begin{subfigure}{0.49\textwidth}
    \includegraphics[width=\textwidth]{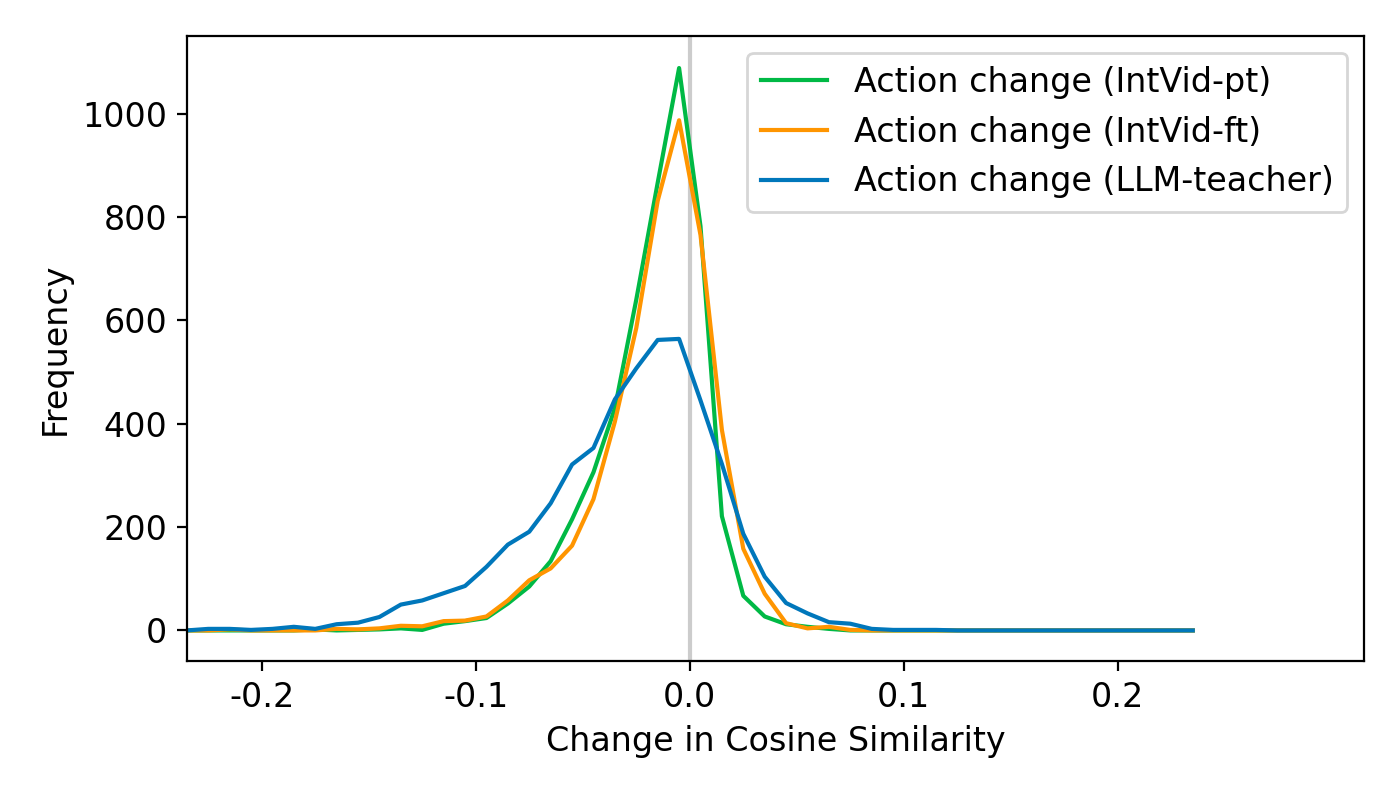}
    \caption{Comparison between $\Delta s$ \wrt action using InternVideo and our \OursMethod.}
    \label{fig:sense_2}
  \end{subfigure}
  \caption{\textbf{Change of cosine similarity \wrt objects or actions.} \textbf{(a):} Comparison between the change in cosine similarity when the action or object is swapped. Results show that current video-text models learn a more effective embedding for objects than for actions. \textbf{(b):} Comparison between the change in cosine similarity using InternVideo or our \OursMethod. This demonstrate that \OursMethod learns a more discriminative embedding for actions by enabling a more effective contrastive learning using knowledge from LLMs. Refer to \cref{sec:main_results} for more details.}
  \label{fig:sense_compare}
\end{figure}

In \cref{fig:sense_compare} we present quantitative results of changes in cosine similarities when the video is unchanged but the object or action in the caption is manipulated. We find that state-of-the-art video-text models learn less discriminative embeddings for actions as compared to objects, which also explains the small gap between image-text and video-text models on standard video-text tasks.

\subsection{\OursMethod} \label{sec:our_method}

We introduce \OursMethod, an LLM-powered approach to learn better action representations with contrastive objectives.
Specifically, LLM serves as the ``teacher'' and provides synthesized captions as extra knowledge for the video-text model to learn from. These new captions serve as negative captions of the original video, removing possible shortcuts and biases and enabling a more effective learning of action semantics.

We start with positive captions from common video-text datasets and run an abstract meaning representation (AMR) parser, resulting in a list of action or object tokens. To generate the hard negative captions, two methods are considered to find novel actions or objects given the original caption and the substitution token $w$.

\subsubsection{Method I: Mask filling.} Masked language modeling (MLM) \cite{devlin2018bert} is a widely adopted self-supervised task for LLMs. In the pretraining stage, a certain percentage of word tokens are substituted with a special token \lstinline{[mask]} and the LLM is optimized to recover the masked tokens. To find negative captions with novel actions given the original caption $c_0$ and substitution token $a$, we first substitute the token $a$ with \lstinline{[mask]} and use a LLM pretrained with MLM to predict a list of $k$ possible tokens $\{\hat{w}_i\}_{i=1}^k$ such that $\hat{w}_i \neq w$ and $\hat{w}_i$ is an action token.

\subsubsection{Method II: LLM-powered chatbot.} Although mask filling can find appropriate words to substitute $w$, it is limited by one single token. In many cases, we want to update the prepositions following the change of verbs. For example, when ``install'' is substituted by ``uninstall'', we should also update the preposition from ``install ... to'' to ``uninstall ... from''. Here we consider a more flexible approach that builds on LLM-powered chatbots. Besides a prompt describing the text substitution, we leverage the in-context learning ability of LLMs \cite{brown2020language,gupta2023visual} and provide multiple in-context examples to obtain the desirable negative captions.

\subsubsection{LLM-teacher.} With the procedure above, we obtain $k$ synthesized negative captions $\{n_i\}_{i=1}^k$ besides the original positive caption $p$ for each video $v$ in existing datasets \cite{xu2016msr,wang2019vatex}. For each of the original and generated captions, LLM serves as a ``teacher'' and provide binary pseudo-labels -- if the caption have matched (``1'') or unmatched (``0'') semantics with the video. Now we may optimize the video-text model with the normalized temperature-scaled cross entropy loss
\begin{align}
    l = - \log \frac{\exp(\text{sim}(f_v, f_p)/\tau)}{\exp(\text{sim}(f_v, f_p) / \tau) + \sum_{i=1}^k \exp(\text{sim}(f_v, f_{n_i}) / \tau)} \label{eq:binary}
\end{align}
where $f_v, f_p, f_{n_i}$ are the visual and textual embeddings.

In practice, we find that certain negative captions have unmatched but similar semantics to the video. It is undesirable to regard them as a strictly negative pair in the contrastive loss. Therefore, we extend the binary pseudo-labels to soft logits by computing caption similarities with a pretrained LLM and then optimize the video-text model to match model outputs with the soft logits from the LLM teacher model \cite{hinton2015distilling}.
\begin{align}
    l & = \mathcal{L}_\text{KL}(z_\text{video-text}, z_\text{LLM}) \label{eq:soft} \\
    z_\text{video-text,t} & = \frac{\exp(\text{sim}(f_v, f_t)/\tau)}{\exp(\text{sim}(f_v, f_p) / \tau) + \sum_{i=1}^k \exp(\text{sim}(f_v, f_{n_i}) / \tau)} \nonumber \\
    z_\text{LLM,t} & = \frac{\exp(\text{sim}(e_p, e_t)/\tau)}{\exp(\text{sim}(e_p, e_p) / \tau) + \sum_{i=1}^k \exp(\text{sim}(e_p, e_{n_i}) / \tau)} \nonumber
\end{align}
where $t \in \{n_i\}_{i=1}^k \cup \{p\}$, $f$'s are textual and visual embeddings of the video-text model, and $e$'s are textual embeddings of a pretrained LLM.

\section{Experiments}

\subsection{Experimental Setup}

\subsubsection{Datasets.} We use MSR-VTT dataset \cite{xu2016msr} and VATEX dataset \cite{wang2019vatex} for standard video-to-text retrieval and our \OursData dataset for retrieval from counterfactually augmented data. Note that \OursData dataset contains the same video sequences for evaluation as in MSR-VTT dataset and VATEX dataset.

\subsubsection{Evaluation metrics.} For standard video-to-text retrieval we report rank-1 accuracy (R@1) as this is the most challenging metric and distinguish the performance of previous video-text models. For retrieval from counterfactually-augmented data we use rank-1 accuracy (R@1), rank-2 accuracy (R@2), and mean rank (MeanR).

\subsubsection{Video-text models.} We consider multiple state-of-the-art public models in this work. CLIP4Clip \cite{luo2022clip4clip} extends CLIP model \cite{radford2021learning} to the video domain using a mean pooling mechanism for zero-shot video-text retrieval and a contrastive loss for finetuning. SimVTP \cite{ma2022simvtp} is pretrained on the WebVid2M dataset \cite{Bain21} with a combination of contrastive learning and masked modeling.
InternVideo \cite{wang2022internvideo} is a video-text foundation model pretrained on a combination of 7 large-scale video-text datasets, with a supervised video post-pretraining for better video recognition. LanguageBind \cite{zhu2024languagebind} is a language-based multi-modal pretraining model with remarkable performance on a wide range of benchmarks.

To demonstrate the efficacy of our \OursMethod, we adopt this method on two state-of-the-art video-text models, SimVTP and InternVideo. In the default setting we generate 10 action-based captions with LLM for each training video. For text generation we use a pretrained XLM-RoBERTa \cite{conneau2019unsupervised} for mask filling. To measure text similarities and compute soft logits, we utilize a pretrained Sentence-BERT model \cite{reimers-2019-sentence-bert}. In \cref{sec:ablation} we ablate on the choices of parameters and caption generation.

\subsection{Main Results} \label{sec:main_results}

\begin{table}[t]
    \centering
    \resizebox{\textwidth}{!}{%
    \begin{tabular}{lccccccccccccc}
        \toprule
        \multirow{2.5}{*}{{Model}} & & \multicolumn{1}{c}{MSR-VTT} & \textcolor{white}{i} & \multicolumn{3}{c}{\OursData (MSR-VTT)} & \textcolor{white}{i} & \multicolumn{1}{c}{VATEX} & \textcolor{white}{i} & \multicolumn{3}{c}{\OursData (VATEX)} \\
        \cmidrule{3-3} \cmidrule{5-7} \cmidrule{9-9} \cmidrule{11-13}
         & & R@1 & & R@1 & R@2 & MeanR & & R@1 & & R@1 & R@2 & MeanR \\
        \midrule
        Human & & & & 95.2 {\scriptsize\textcolor{white}{$\downarrow$1.1}} & & & & & & 96.8 {\scriptsize\textcolor{white}{$\downarrow$1.1}} & & \\
        Random & & <1e-3 {\scriptsize\textcolor{white}{$\downarrow$1.1}} & & 16.7 {\scriptsize\textcolor{white}{$\downarrow$1.1}} & & & & <1e-3 {\scriptsize\textcolor{white}{$\downarrow$1.1}} & & 16.7 {\scriptsize\textcolor{white}{$\downarrow$1.1}} & & \\
        \midrule \midrule
        \textbf{\textit{Zero-Shot}} \\
        CLIP \cite{radford2021learning} & & 26.3 {\scriptsize\textcolor{white}{$\downarrow$1.1}} & & 37.3 {\scriptsize\textcolor{white}{$\downarrow$1.1}} & 55.3 {\scriptsize\textcolor{white}{$\downarrow$1.1}} & 2.6 {\scriptsize\textcolor{white}{$\downarrow$1.1}} & & 38.8 {\scriptsize\textcolor{white}{$\downarrow$1.1}} & & 34.8 {\scriptsize\textcolor{white}{$\downarrow$1.1}} & 54.3 {\scriptsize\textcolor{white}{$\downarrow$1.1}} & 2.7 {\scriptsize\textcolor{white}{$\downarrow$1.1}} \\
        VideoCLIP \cite{xu-etal-2021-videoclip} & & 14.5 {\scriptsize\textcolor{white}{$\downarrow$1.1}} & & 35.1 {\scriptsize\textcolor{white}{$\downarrow$1.1}} & 71.3 {\scriptsize\textcolor{white}{$\downarrow$1.1}} & 2.6 {\scriptsize\textcolor{white}{$\downarrow$1.1}} & & 13.7 {\scriptsize\textcolor{white}{$\downarrow$1.1}} & & 33.0 {\scriptsize\textcolor{white}{$\downarrow$1.1}} & 70.7 {\scriptsize\textcolor{white}{$\downarrow$1.1}} & 2.7 {\scriptsize\textcolor{white}{$\downarrow$1.1}} \\
        InternVideo \cite{wang2022internvideo} & & 37.5 {\scriptsize\textcolor{white}{$\downarrow$1.1}} & & 45.8 {\scriptsize\textcolor{white}{$\downarrow$1.1}} & 63.6 {\scriptsize\textcolor{white}{$\downarrow$1.1}} & 2.3 {\scriptsize\textcolor{white}{$\downarrow$1.1}} & & 76.9 {\scriptsize\textcolor{white}{$\downarrow$1.1}} & & 44.1 {\scriptsize\textcolor{white}{$\downarrow$1.1}} & 63.9 {\scriptsize\textcolor{white}{$\downarrow$1.1}} & 2.3 {\scriptsize\textcolor{white}{$\downarrow$1.1}} \\
        LanguageBind \cite{zhu2024languagebind} & & 42.8 {\scriptsize\textcolor{white}{$\downarrow$1.1}} & & 41.3 {\scriptsize\textcolor{white}{$\downarrow$1.1}} & 77.0 {\scriptsize\textcolor{white}{$\downarrow$1.1}} & 2.4 {\scriptsize\textcolor{white}{$\downarrow$1.1}} & & \textcolor{white}{00.0} {\scriptsize\textcolor{white}{$\downarrow$1.1}} & & 42.3 {\scriptsize\textcolor{white}{$\downarrow$1.1}} & 77.0 {\scriptsize\textcolor{white}{$\downarrow$1.1}} & 2.4 {\scriptsize\textcolor{white}{$\downarrow$1.1}} \\
        \midrule \midrule
        \textbf{\textit{Finetuned}} \\
        LanguageBind \cite{zhu2024languagebind} \\
        CLIP4Clip \cite{luo2022clip4clip} & & 43.1 {\scriptsize\textcolor{white}{$\downarrow$1.1}} & & 50.8 {\scriptsize\textcolor{white}{$\downarrow$1.1}} & 72.4 {\scriptsize\textcolor{white}{$\downarrow$1.1}} & 2.0 {\scriptsize\textcolor{white}{$\downarrow$1.1}} \\
        VindLU \cite{cheng2022vindlu} & & 46.6 {\scriptsize\textcolor{white}{$\downarrow$1.1}} & & 53.4 {\scriptsize\textcolor{white}{$\downarrow$1.1}} & 70.9 {\scriptsize\textcolor{white}{$\downarrow$1.1}} & 2.0 {\scriptsize\textcolor{white}{$\downarrow$1.1}} \\
        SimVTP \cite{ma2022simvtp} & & \textbf{50.2} {\scriptsize\textcolor{white}{$\downarrow$1.1}} & & 35.7 {\scriptsize\textcolor{white}{$\downarrow$1.1}} & 70.8 {\scriptsize\textcolor{white}{$\downarrow$1.1}} & 2.6 {\scriptsize\textcolor{white}{$\downarrow$1.1}} & & 76.6 {\scriptsize\textcolor{white}{$\downarrow$1.1}} & & 33.6 {\scriptsize\textcolor{white}{$\downarrow$1.1}} & 68.4 {\scriptsize\textcolor{white}{$\downarrow$1.1}} & 2.6 {\scriptsize\textcolor{white}{$\downarrow$1.1}} \\
        \cellcolor{mygray-bg}{\emph{\ w/} \OursMethod-lbl} & \cellcolor{mygray-bg}{} & \cellcolor{mygray-bg}{49.0 {\scriptsize\textcolor{ablation_red}{$\downarrow$1.2}}} & \cellcolor{mygray-bg}{} & \cellcolor{mygray-bg}{40.0 {\scriptsize\textcolor{ablation_green}{$\uparrow$4.3}}} & \cellcolor{mygray-bg}{73.1 {\scriptsize\textcolor{ablation_green}{$\uparrow$2.3}}} & \cellcolor{mygray-bg}{2.3 {\scriptsize\textcolor{ablation_green}{$\downarrow$0.3}}} & \cellcolor{mygray-bg}{} & \cellcolor{mygray-bg}{74.8 {\scriptsize\textcolor{ablation_red}{$\downarrow$1.8}}} & \cellcolor{mygray-bg}{} & \cellcolor{mygray-bg}{37.3 {\scriptsize\textcolor{ablation_green}{$\uparrow$3.7}}} & \cellcolor{mygray-bg}{72.2 {\scriptsize\textcolor{ablation_green}{$\uparrow$3.8}}} & \cellcolor{mygray-bg}{2.4 {\scriptsize\textcolor{ablation_green}{$\downarrow$0.2}}} \\
        \cellcolor{mygray-bg}{\emph{\ w/} \OursMethod-lgt} & \cellcolor{mygray-bg}{} & \cellcolor{mygray-bg}{\underline{49.5} {\scriptsize\textcolor{ablation_red}{$\downarrow$0.7}}} & \cellcolor{mygray-bg}{} & \cellcolor{mygray-bg}{43.5 {\scriptsize\textcolor{ablation_green}{$\uparrow$7.8}}} & \cellcolor{mygray-bg}{75.0 {\scriptsize\textcolor{ablation_green}{$\uparrow$4.2}}} & \cellcolor{mygray-bg}{2.2 {\scriptsize\textcolor{ablation_green}{$\downarrow$0.4}}} & \cellcolor{mygray-bg}{} & \cellcolor{mygray-bg}{75.3 {\scriptsize\textcolor{ablation_red}{$\downarrow$1.3}}} & \cellcolor{mygray-bg}{} & \cellcolor{mygray-bg}{40.1 {\scriptsize\textcolor{ablation_green}{$\uparrow$6.5}}} & \cellcolor{mygray-bg}{73.5 {\scriptsize\textcolor{ablation_green}{$\uparrow$5.1}}} & \cellcolor{mygray-bg}{2.3 {\scriptsize\textcolor{ablation_green}{$\downarrow$0.3}}} \\
        InternVideo \cite{wang2022internvideo} & & 49.1 {\scriptsize\textcolor{white}{$\downarrow$1.1}} & & 58.6 {\scriptsize\textcolor{white}{$\downarrow$1.1}} & 80.2 {\scriptsize\textcolor{white}{$\downarrow$1.1}} & 1.8 {\scriptsize\textcolor{white}{$\downarrow$1.1}} & & \textbf{87.9} {\scriptsize\textcolor{white}{$\downarrow$1.1}} & & 58.2 {\scriptsize\textcolor{white}{$\downarrow$1.1}} & 76.9 {\scriptsize\textcolor{white}{$\downarrow$1.1}} & 1.9 {\scriptsize\textcolor{white}{$\downarrow$1.1}} \\
        \cellcolor{mygray-bg}{\emph{\ w/} \OursMethod-lbl} & \cellcolor{mygray-bg}{} & \cellcolor{mygray-bg}{48.2 {\scriptsize\textcolor{ablation_red}{$\downarrow$0.9}}} & \cellcolor{mygray-bg}{} & \cellcolor{mygray-bg}{64.2 {\scriptsize\textcolor{ablation_green}{$\uparrow$5.6}}} & \cellcolor{mygray-bg}{82.5 {\scriptsize\textcolor{ablation_green}{$\uparrow$2.3}}} & \cellcolor{mygray-bg}{1.7 {\scriptsize\textcolor{ablation_green}{$\downarrow$0.1}}} & \cellcolor{mygray-bg}{} & \cellcolor{mygray-bg}{85.2 {\scriptsize\textcolor{ablation_red}{$\downarrow$2.7}}} & \cellcolor{mygray-bg}{} & \cellcolor{mygray-bg}{63.8 {\scriptsize\textcolor{ablation_green}{$\uparrow$5.6}}} & \cellcolor{mygray-bg}{80.5 {\scriptsize\textcolor{ablation_green}{$\uparrow$3.6}}} & \cellcolor{mygray-bg}{1.7 {\scriptsize\textcolor{ablation_green}{$\downarrow$0.2}}} \\
        \cellcolor{mygray-bg}{\emph{\ w/} \OursMethod-lgt} & \cellcolor{mygray-bg}{} & \cellcolor{mygray-bg}{48.9 {\scriptsize\textcolor{ablation_red}{$\downarrow$0.2}}} & \cellcolor{mygray-bg}{} & \cellcolor{mygray-bg}{\textbf{65.8} {\scriptsize\textcolor{ablation_green}{$\uparrow$7.2}}} & \cellcolor{mygray-bg}{\textbf{83.8} {\scriptsize\textcolor{ablation_green}{$\uparrow$3.6}}} & \cellcolor{mygray-bg}{\textbf{1.7} {\scriptsize\textcolor{ablation_green}{$\downarrow$0.1}}} & \cellcolor{mygray-bg}{} & \cellcolor{mygray-bg}{\underline{87.3} {\scriptsize\textcolor{ablation_red}{$\downarrow$0.6}}} & \cellcolor{mygray-bg}{} & \cellcolor{mygray-bg}{\textbf{65.6} {\scriptsize\textcolor{ablation_green}{$\uparrow$7.4}}} & \cellcolor{mygray-bg}{\textbf{81.7} {\scriptsize\textcolor{ablation_green}{$\uparrow$4.8}}} & \cellcolor{mygray-bg}{\textbf{1.7} {\scriptsize\textcolor{ablation_green}{$\downarrow$0.2}}} \\
        \bottomrule
    \end{tabular}}
    \caption{Performance of standard video-to-text retrieval on MSR-VTT dataset \cite{xu2016msr} and VATEX dataset \cite{wang2019vatex}, as compared to performance of retrieval from counterfactually-augmented data on our \OursData dataset. In the ``zero-shot'' setting, models are evaluated directly after the pretraining stage, while in the ``finetuned'' setting, models are finetuned on the training set of MSR-VTT or VATEX. With \OursMethod, we enable a more efficient learning of action semantics and effectively improves R@1 accuracy on \OursData. Here ``\OursMethod-lbl'' stands for our approach with binary pseudo-labels and ``\OursMethod-lgt'' uses soft logits.}
    \label{tab:main_results}
\end{table}

\subsubsection{Performance of previous state-of-the-art.} In \cref{tab:main_results} we report the quantitative results of standard video-text retrieval on MSR-VTT and VATEX, and of retrieval from counterfactually augmented data on our \OursData. We also report a ``Random'' performance where a model predicts random guesses. We make the following observations: (i) Results show that previous state-of-the-art video-text models demonstrate limited understanding of the action semantics in a video with less than 60\% R@1 accuracy on RCAD, given a 16.7\% R@1 accuracy when taking random guesses. (ii) Although it is hard to compare the quantitative results between two tasks, we note that our RCAD is much more challenging given the performance gap between ``Random'' and previous state-of-the-arts. This supports our previous arguments that our RCAD removes shortcuts from standard video-text datasets and focuses on harder questions in the video domain that require cross-frame reasoning.

\subsubsection{Performance of \OursMethod.} We apply our \OursMethod to two pretrained video-text models, SimVTP \cite{ma2022simvtp} and InternVideo \cite{wang2022internvideo}. Specifically, we consider the two objectives in \cref{eq:binary} and \cref{eq:soft}, labeled as ``\OursMethod-lbl'' and ``\OursMethod-lgt'' respectively. Results in \cref{tab:main_results} show that by exploiting knowledge from pretrained LLMs, \OursMethod enables a more efficient leaning of action semantics and achieved improved results on various metrics of RCAD, increasing the R@1 accuracy on RCAD by 7.2\% and 7.4\%. We also note that with binary pseudo-labels, results demonstrate a trade-off between the performance on standard video-to-text retrieval and RCAD; and in comparison, \OursMethod with soft logits achieves the highest performance on RCAD (65.8\% and 65.6\%) with negligible drops (0.2\% and 0.6\%) on standard retrieval. This is because soft logits account for the text similarities between different actions and avoid overfitting on binary pseudo-labels.

\subsubsection{Comparison with human-level performance.} The human-level performance are reported in \cref{tab:main_results}. We see that human annotators achieved an almost perfect performance on \OursData dataset, with a 95.2\% accuracy on MSR-VTT videos and a 96.8\% accuracy on VATEX videos. These results show that: (i) most questions for retrieval from counterfactually-augmented data are answerable with an unique answer, and (ii) this task is fairly simple for human. As a comparison, all previous video-text models with large-scale pretraining fall behind by a wide margin. Specifically, the state-of-the-art video-text model InternVideo \cite{wang2022internvideo} features a heavy two-stage pretraining on a collection of 7 large-scale video-text datasets but falls behind by 36.6\% and 38.6\% even with downstream finetuning.

\subsubsection{Analyses of failure cases.} In \cref{fig:teaser_compare} we present two failure cases by the InternVideo \cite{wang2022internvideo} model on our \OursData dataset. Both examples are trivial for human, yet challenging from video-text models due to incapability of cross-frame reasoning. For the first example, a model must analyze the interactions between the person and the rock over a sequence of frames to predict the correct answer. And in the second example, the model must observe the change of appearances over time. As we can see, with the counterfactually augmented data in our \OursData, models cannot ``guess'' the correct answer by exploiting shortcuts extracted from a single frame. The findings on \OursData highlight significant weaknesses in current video-text models, offering valuable insights for future researches in this area. Additional qualitative examples of video-text models on \OursData dataset are provided in the supplementary materials for readers' reference.

\subsubsection{Sensitivity to the change of object or action.} With the counterfactually augmented data collected, we investigate how the cosine similarities between the video and text change when the object or action in the text are swapped with an object or action not present in the image. Given the original video-text pair $(v, t)$ and a counterfactually-augmented text $\hat{t}$, the \textit{change in cosine similarity} is given by $\Delta s = s(v, \hat{t}) - s(v, t)$. As $\hat{t}$ contains objects or actions not present in the video, $\Delta s$ should be negative for an ideal video-text model. Moreover, larger $\lvert \Delta s \rvert$ implies the model being more sensitive to the changes. We compute $\Delta s$ using the InternVideo with pretraining only (``IntVid-pt''), InternVideo with downstream finetuning (``IntVid-ft''), and our \OursMethod (``\OursMethod'').

Results in \cref{fig:sense_1} show that the $\Delta s$ are almost always negative when the objects are swapped and the absolute changes are larger. In comparison, $\Delta s$ are sometimes positive when actions are modified and the absolute changes are smaller. This demonstrate that InternVideo learns a more effective embedding for objects than for actions. This further highlights the necessity of our new evaluation paradigm for exploring the limitations of video-text models beyond existing tasks. Moreover, in \cref{fig:sense_2} we note that with \OursMethod the action embeddings are more effective than the ones learned by InternVideo, as shown by the more reasonable changes in cosine similarity when actions are swapped.

Please refer to the supplementary materials for a detailed discussion regarding the influence of textual encoders.

\subsection{Ablation Study} \label{sec:ablation}

We conduct ablation studies on the VATEX dataset \cite{wang2019vatex} and our \OursData dataset. We follow the same settings as above and report rank-1 (R@1) accuracy for standard video-to-text retrieval and rank-1 accuracy (R@1), rank-2 accuracy (R@2), and mean rank (MeanR) for counterfactual augmented data.

\subsubsection{Choice of captions in \OursMethod.} \cref{tab:ablation} shows the comparison between different numbers and types of LLM captions generated. In the ``default'' setting we use 10 action-based LLM captions and compare to settings with 5 action-based LLM captions or 5 action-based and 5 object-based LLM captions. Results show with 5 action-based LLM captions the R@1 accuracy on \OursData drops by 0.9\% while the R@1 accuracy on VATEX increases by 0.3\%. We also experiment on object-based LLM captions and find them not beneficial for both standard video-to-text retrieval or our RCAD. This is consistent with our assumption that pretrained video-text models already learns a discriminative embedding for objects and a more efficient training objective for action is needed.

\subsubsection{Choice of caption generation.} Besides a pretrained XLM-RoBERTa model for mask filling, we also experiment on generating captions with LLM-powered agents that are finetuned on chat datasets for dialogue applications.
Specifically we use the ``chat'' model finetuned for dialogue applications.
Empirically we find that captions generated by LLM chatbots achieves a higher overall quality, exploring a more diverse caption space and better flexibility. However we don't observe significant improvements from our ablation study experiments -- the R@1 accuracy on VATEX drops by 0.3\% and the R@1 accuracy on \OursData increases by 0.3\%. We choose XLM-RoBERTa in our main experiments as the model runs faster and scale up easily to bigger settings. We provide qualitative comparisons between the two types of caption generation in our supplementary materials.

\begin{table}[t]
    \centering
    \resizebox{0.75\textwidth}{!}{%
    \begin{tabular}{lccccccccccccc}
        \toprule
        \multirow{2.5}{*}{{Model}} & & \multicolumn{1}{c}{VATEX} & \textcolor{white}{i} & \multicolumn{3}{c}{\OursData (VATEX)} \\
        \cmidrule{3-3} \cmidrule{5-7}
         & & R@1 & & R@1 & R@2 & MeanR \\
        \midrule
        Default & & 87.3 {\scriptsize\textcolor{white}{$\downarrow$0.9}} & & 65.6 {\scriptsize\textcolor{white}{$\downarrow$0.9}} & 81.7 {\scriptsize\textcolor{white}{$\downarrow$0.9}} & 1.7 {\scriptsize\textcolor{white}{$\downarrow$0.9}} \\
        \textit{(10 action captions; XLM-RoBERTa)} \\
        \midrule
        5 action captions & & 87.6 {\scriptsize\textcolor{ablation_green}{$\uparrow$0.3}} & & 64.7 {\scriptsize\textcolor{ablation_red}{$\downarrow$0.9}} & 81.0 {\scriptsize\textcolor{ablation_red}{$\downarrow$0.7}} & 1.7 {\scriptsize\textcolor{white}{$\uparrow$}\textcolor{mygray}{0.0}} \\
        5 object + 5 object captions & & 87.5 {\scriptsize\textcolor{ablation_green}{$\uparrow$0.2}} & & 64.2 {\scriptsize\textcolor{ablation_red}{$\downarrow$1.4}} & 80.6 {\scriptsize\textcolor{ablation_red}{$\downarrow$1.1}} & 1.7 {\scriptsize\textcolor{white}{$\uparrow$}\textcolor{mygray}{0.0}} \\
        LLM Chatbot & & 87.0 {\scriptsize\textcolor{ablation_red}{$\downarrow$0.3}} & & 65.9 {\scriptsize\textcolor{ablation_green}{$\uparrow$0.3}} & 81.8 {\scriptsize\textcolor{ablation_green}{$\uparrow$0.1}} & 1.7 {\scriptsize\textcolor{white}{$\uparrow$}\textcolor{mygray}{0.0}} \\
        \bottomrule
    \end{tabular}%
    }
    \caption{Ablation studies on the number of LLM captions and caption generation methods. In the default setting we use a total of 10 action-based LLM captions and utilize a pretrained XLM-RoBERTa \cite{conneau2019unsupervised} for mask filling.}
    \label{tab:ablation}
\end{table}

\section{Conclusions}

In this work we propose a new evaluation task, retrieval from counterfactually augmented data, and a benchmark dataset, \OursData. The idea is to remove shortcuts in the video-text questions with a human-in-the-loop system and produce more challenging questions where the video-text model must derive a comprehensive understanding of the video with cross-frame reasoning. Quantitative and qualitative evaluation results on our \OursData dataset show that despite the task is trivial for human, previous state-of-the-art video-text models can be easily fooled by the counterfactually augmented data. This implies that the prominent results on previous video-text benchmarks could be misleading -- models may largely exploit shortcuts without genuinely understand the video contents. Moreover, we identify a key limitation of current contrastive learning on video-text data being the shortcut learning and propose LLM-teacher that enables a more effective learning of action semantics by utilizing knowledge from pretrained LLMs. Experimental results and analyses show that our method can learn a more discriminative representation for actions.


\subsubsection{Supplementary materials.} We present the following: (1)~limitations of our work, (2)~more details about our \OursData dataset, (3)~ethics statements, and (4)~extra quantitative and qualitative results.

\subsubsection{Acknowledgements.} We would like to thank Daniel Khashabi, Yiyan Li, and the anonymous reviewers for their helpful comments and suggestions. Wufei Ma is supported by ONR with N00014-23-1-2641.

%
\bibliographystyle{splncs04}
\bibliography{main}


\newpage

\renewcommand{\thesection}{\Alph{section}}
\setcounter{section}{0}

\section{\OursData: Data Collection}

\subsection{Data Collection}

\begin{figure}[t]
  \centering
  \includegraphics[width=0.9\textwidth]{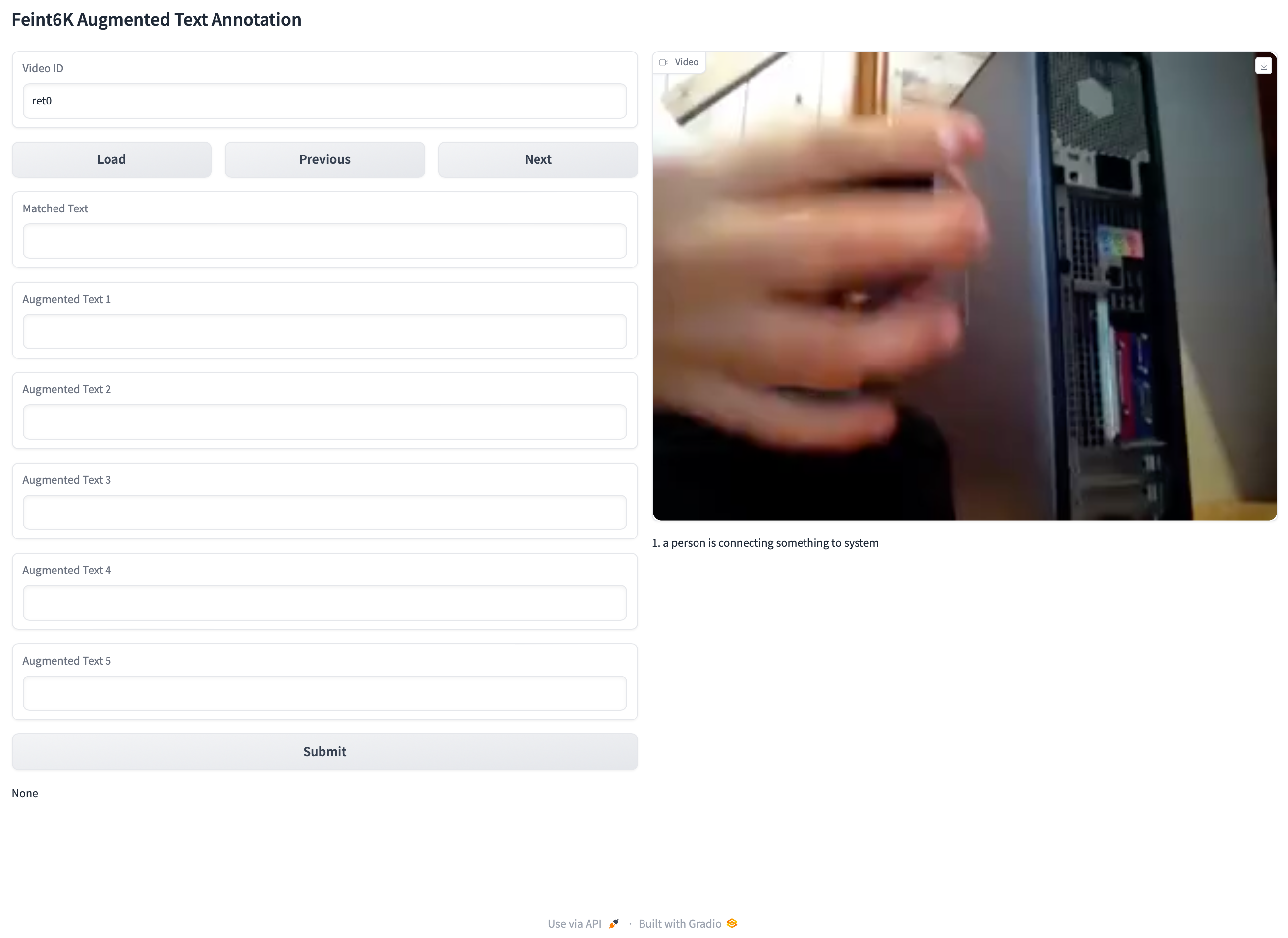}
  \caption{Screenshot of the Gradio app we develop for caption collection.}
  \label{fig:supp_app}
\end{figure}

We develop a Gradio app to collect counterfactually augmented captions from the annotators. For each question, the annotator is first presented with the video and caption(s) from the MSR-VTT \cite{xu2016msr} or VATEX \cite{wang2019vatex} dataset. Then the annotator is asked to provide a groundtruth text with the matched action in the video, and another five counterfactually augmented texts with novel actions not present in the video. A screenshot of our augmented text annotation web app is demonstrated in \cref{fig:supp_app}.

\subsection{Dataset Statistics} \label{sec:supp_stats}

We annotate a total of 6,243 videos, each with 5 counterfactually augmented captions. Original videos come from the validation set of the MSR-VTT dataset \cite{xu2016msr} (947 videos) and the test set of the VATEX dataset \cite{wang2019vatex} (5296 videos).

\subsection{Annotation Guidelines}

To get annotators familiar with the high-level task, \ie, video-text understanding, and the specific work to accomplish, \ie, annotating counterfactually augmented texts, we provide detailed guidelines about the goal of this project, expected outcomes from the annotations, good and bad practice, \etc. The full annotation guidelines is available from our project page.

\subsection{Notes}

\subsubsection{Certain actions appear more frequently than others.} Although we encourage annotators to design diverse actions that are plausible given the context in the video, annotators may sometimes resort to simpler and more common actions when novel actions are hard to come up with. Actions such as ``jump'', ``climb'', and ``laugh'' appear more frequently than other actions in our annotated counterfactually augmented captions.

\subsubsection{Actions in \OursData explore a much broader space of actions than standard datasets.} Actions in standard video-text datasets are limited to common videos available from the internet. As annotators are freely exploring open-set actions that fits the context in the video, actions in \OursData could explore a much broader space of actions, such as ``kick a snowman'', ``throw up a guitar'', or ``drop a watermelon''.

\subsection{Ethics}

The data we collect from the human annotators include the counterfactually augmented captions as well as categorical prediction labels when evaluating human performance. Before starting the annotation work, each annotator first sign the consent form acknowledging that: (i) they choose to participate this program voluntarily; (ii) the collected annotations will be used in video-text research projects; and (iii) the collected annotations will be open-sourced and shared with other research groups.

\section{Limitations}

Although we attempt to remove shortcuts from current video-text datasets for a better evaluation of video-text understanding, our evaluation task RCAD is still limited by certain biases in these datasets. As all testing videos are sampled from web-collected data, the object-action pairs would follow a long-tail distribution and video-text models may exploit these biases for a higher benchmark performance. For future work we consider using pretrained text-to-video generation models or video editing models to address the bias issue.

\section{Shortcuts in Multi-Modal Contrastive Learning} \label{sec:textenc}

\begin{figure}[t]
  \centering
  \includegraphics[width=\textwidth]{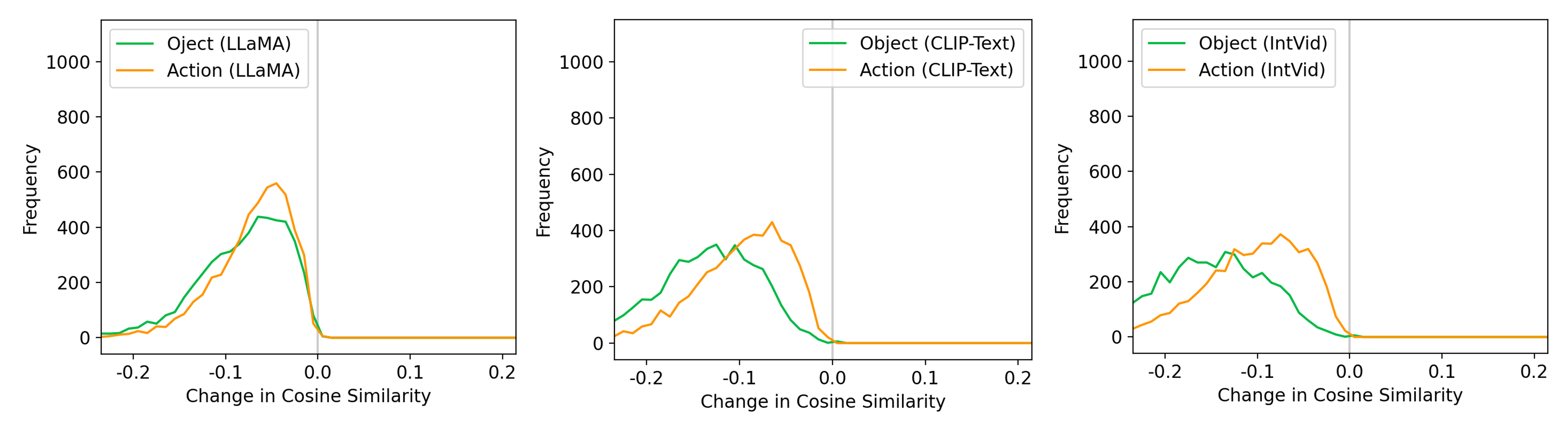}
  \caption{Comparing changes in cosine similarities using different textual encoders. See discussions in \cref{sec:textenc}.}
  \label{fig:textenc}
\end{figure}

In \cref{sec:method_shortcuts} we compute the change of cosine similarities when videos are unchanged but objects or actions in the captions are manipulated. Results in \cref{fig:sense_1} demonstrate very different patterns for changes in objects and actions. This gap is attributed to: (i) vision encoder's inability to distinguish between different actions from cross-frame reasoning, and (ii) sensitivity of textual embeddings \textit{w.r.t.} various syntactic categories. To investigate the influence from the textual encoders, we visualize changes in cosine similarities between textual embeddings obtained from (1) LLM2Vec \cite{behnamghader2024llm2vec}, a textual encoder finetuned from LLaMA without multi-modal training, (ii) pretrained textual encoder used in InternVideo, \textit{i.e.}, CLIP textual encoder, and (iii) textual encoder from InternVideo. Results in \cref{fig:textenc} show that there is a significant gap between LLM textual encoders and encoders trained with multi-modal contrastive learning. This supports our discussions about shortcuts in \cref{sec:method_shortcuts}, where objects become shortcuts in contrastive learning and hinders the models from learning effective action representations.

\section{Qualitative Comparisons of Caption Generation}

\begin{figure}[t]
  \centering
  \includegraphics[width=0.9\textwidth]{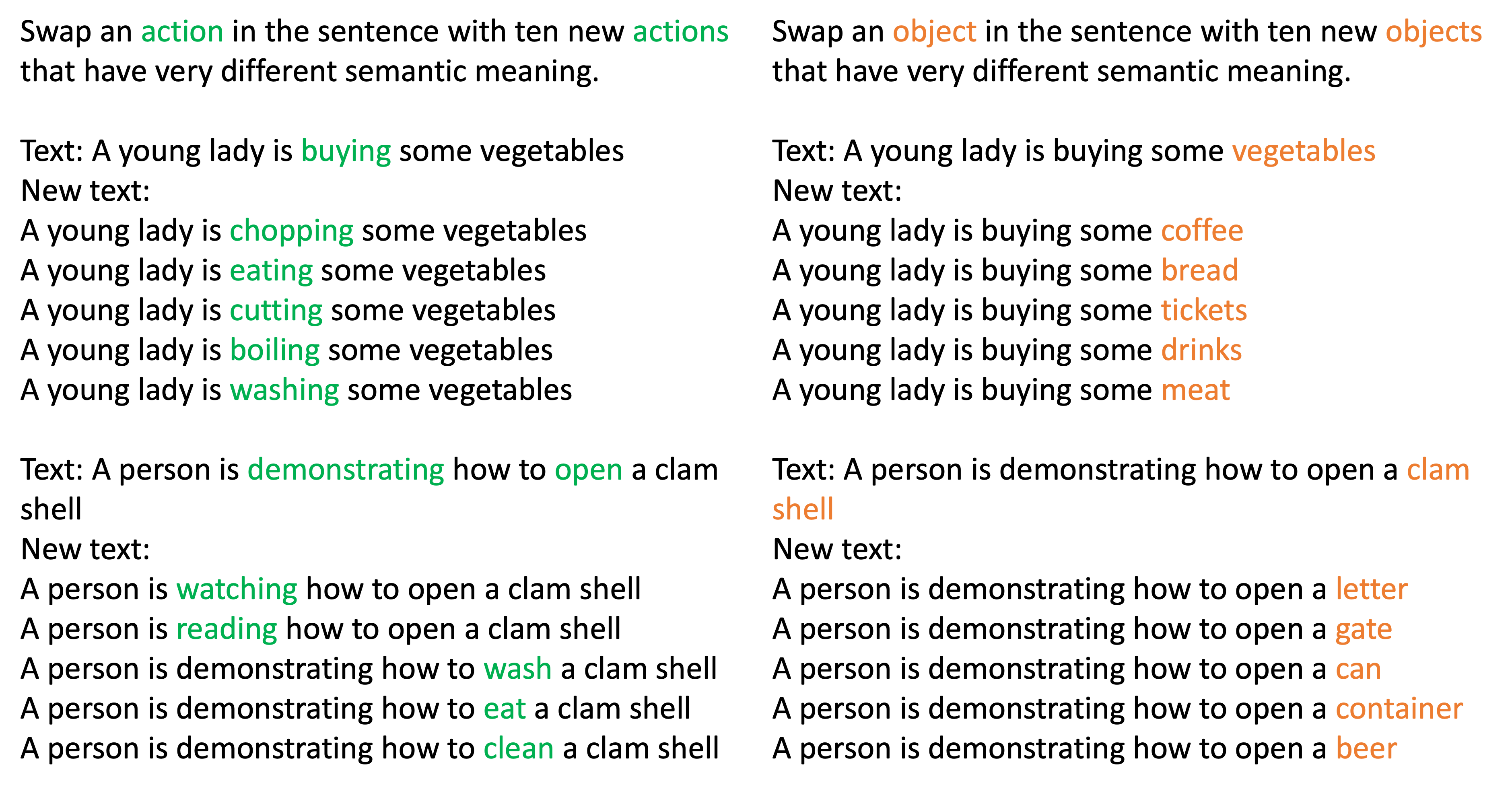}
  \caption{We leverage the in-context learning capabilities of LLMs and obtain desirable captions with a LLM-powered chatbot.}
  \label{fig:incontext}
\end{figure}

\begin{figure}[t]
  \centering
  \includegraphics[width=\textwidth]{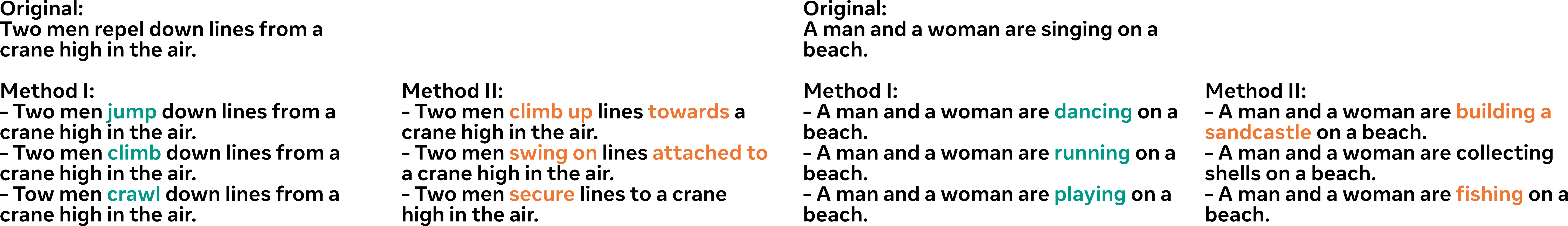}
  \caption{Qualitative comparisons between captions generated by Method I and Method II. In general we find Method II is capable of exploring a more diverse space of caption and performing multi-word substitutions.}
  \label{fig:compare_captiongen}
\end{figure}

As detailed in \cref{sec:our_method}, we consider two caption generation methods in our \OursMethod to obtain ``hard'' negative captions for the models to learn from. In Method I, we used a pretrained XLM-RoBERTa \cite{conneau2019unsupervised} model to perform mask filling. In Method II, we leverage the in-context learning capabilities of LLMs and obtain desirable captions with a LLM-powered chatbot (see \cref{fig:incontext}). In \cref{fig:compare_captiongen} we present some qualitative comparisons between captions generated by Method I and II. We find that Method I relies heavily on post-processing such as rule-based filtering (\eg, removing ambiguous words -- ``he'' and ``it'') or language-based filtering (\eg, removing repeated words derived from the same root -- ``merged'' and ``merging''). Meanwhile, Method II is also capable of updating the preposition words according to the change of actions or generating actions composed of multiple words.

\section{Qualitative Results on \OursData}

\begin{figure}[t]
  \centering
  \includegraphics[width=\textwidth]{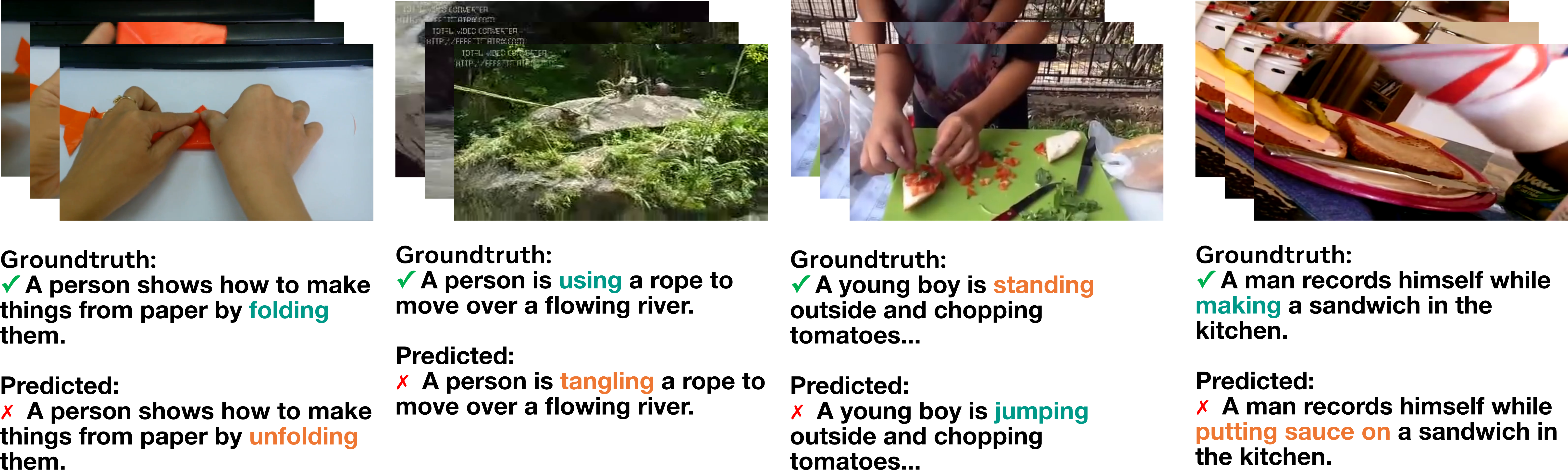}
  \caption{Failure cases of InternVideo \cite{wang2022internvideo} on RCAD in our \OursData dataset. We show that RCAD is hard for video-text models as it requires complex cross-frame reasoning that current models are weak at.}
  \label{fig:failure_cases}
\end{figure}

In \cref{fig:failure_cases} we present some additional failure cases of InternVideo \cite{wang2022internvideo} on retrieval from counterfactually augmented data in our \OursData dataset. Despite the task being trivial for humans, we show that RCAD is hard for video-text models as it requires complex cross-frame reasoning that current models are weak at. In the ``folding/unfolding'' example, the model must perceive and reason about changes of the paper over time. In the ``using/tangling'' example, the model must reason about the interactions between the ``person'' and the ``rope'' over a sequence of frames.

The evaluation results we present on RCAD demonstrate that current video-text models still fall far behind human-level understanding of videos and calls for more advanced pretraining strategies. Our \OursMethod introduces a more effective contrastive learning objective and presents an early step towards this goal.

\subsubsection{Improvements from LLM-teacher.} 
\begin{figure}[t]
  \centering
  \includegraphics[width=0.5\textwidth]{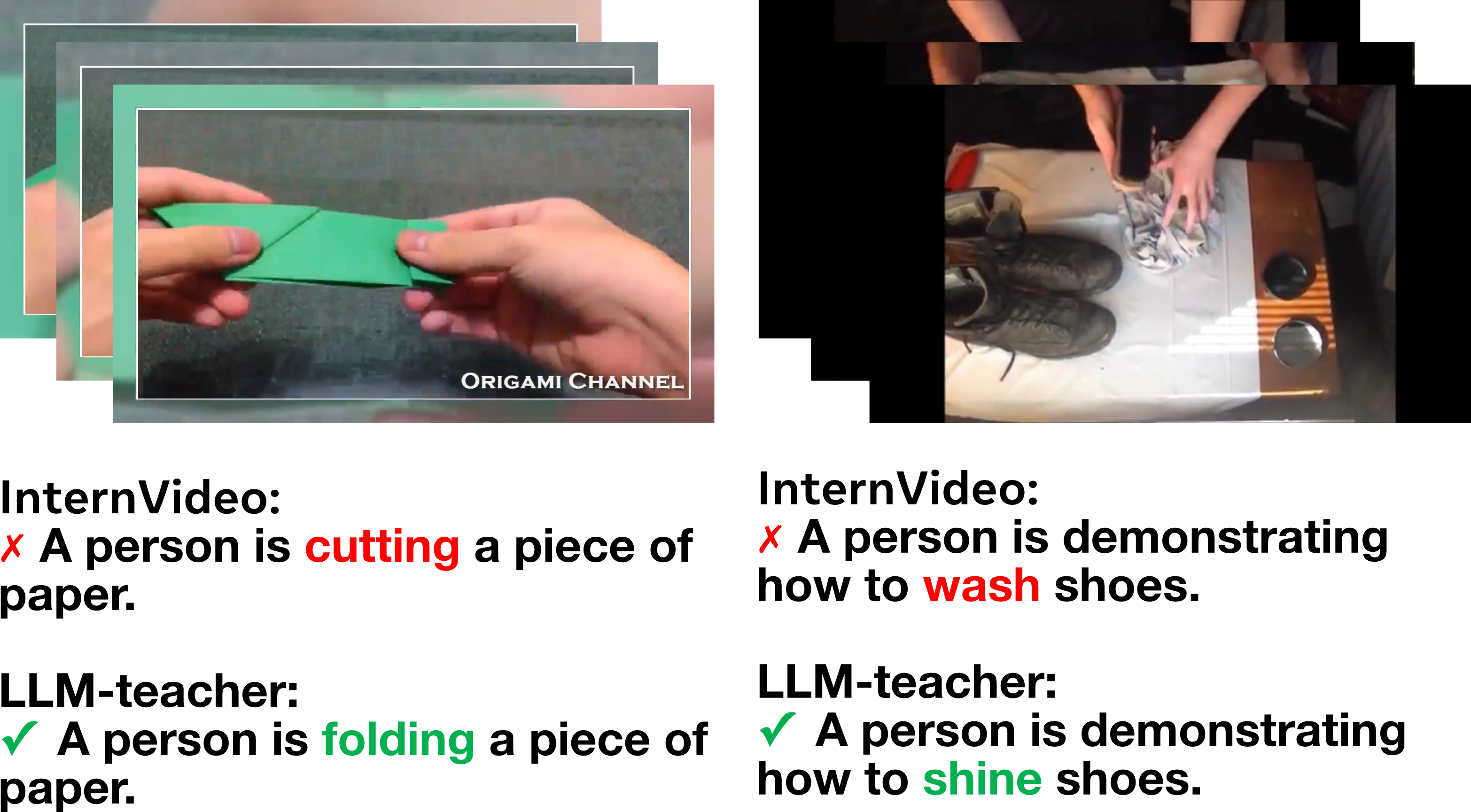}
  \caption{We analyze the improvements of LLM-teacher as compared to InternVideo \cite{wang2022internvideo} baseline. We find that with LLM-teacher, our model learns to distinguish actions more effectively.}
  \label{fig:llm_teacher}
\end{figure}

We analyze the improvements of LLM-teacher as compared to the InternVideo \cite{wang2022internvideo} baseline. We find that with LLM-teacher, our model learns to distinguish actions more effectively (see \cref{fig:llm_teacher}). However, it still struggles to understand complex human activities (\textit{e.g.}, ``jump or crawl on a bounce house'') and fine-grained activities (\textit{e.g.}, ``apply or throw a lipstick'').

\newpage

\end{document}